\documentclass{article}
\usepackage{spconf,amsmath,graphicx}
\usepackage{subcaption}
\usepackage{booktabs}
\usepackage{adjustbox}
\usepackage{multirow}
\usepackage{bbding}
\usepackage{multirow}
\usepackage[table]{xcolor}
\usepackage[hyphens]{url}
\usepackage{amsmath}
\usepackage{amsfonts}
\usepackage{graphicx}
\usepackage{scrextend}

\usepackage[colorlinks=true,urlcolor=blue,linkcolor=red,citecolor=blue]{hyperref}
\usepackage{arydshln}

\usepackage{acronym}
\newacro{gan}[GAN]{generative adversarial network}
\newacro{dire}[DIRE]{diffusion reconstruction error}
\newacro{cnn}[CNN]{convolution neural network}
\newacro{vlm}[VLM]{vision-language model}
\newacro{vit}[ViT]{vision transformer}
\newacro{nlp}[NLP]{natural language processing}
\newacro{cv}[CV]{computer vision}
\newacro{blip}[BLIP]{bootstrapping language image pre-training}
\newacro{vqa}[VQA]{visual question answering}
\newacro{lora}[LoRA]{low-rank adaptation}
\newacro{peft}[PEFT]{parameter-efficient fine-tuning}
\newacro{gpt}[GPT]{generative pre-trained transformer}
\newacro{q-former}[Q-Former]{querying transformer}
\newacro{sedid}[SeDID]{stepwise error for diffusion-generated image detection}
\newacro{sd}[SD]{stable diffusion}
\newacro{lsun}[LSUN]{large-scale scene understanding}
\newacro{fc}[FC]{fully-connected}
\newacro{deit}[DeiT]{data-efficient image transformers}
\newacro{ldm}[LDM]{latent diffusion model}

\title{Harnessing the Power of Large Vision Language Models for Synthetic Image Detection}
\name{Mamadou Keita\textsuperscript{1},
Wassim Hamidouche\textsuperscript{2},
Hassen Bougueffa\textsuperscript{1},
Abdenour Hadid\textsuperscript{3},
Abdelmalik Taleb-Ahmed\textsuperscript{1}}
\address{ \textsuperscript{1} Univ. Polytechnique Hauts-de-France, France \\
 \textsuperscript{2} Technology Innovation Institute P.O.Box: 9639,
Masdar City, Abu Dhabi, UAE\\
 \textsuperscript{3} Sorbonne Center for Artificial Intelligence, Sorbonne University Abu Dhabi, UAE
}
\begin{document}
\ninept
\maketitle
\begin{abstract}
In recent years, the emergence of models capable of generating images from text has attracted considerable interest, offering the possibility of creating realistic images from text descriptions. Yet these advances have also raised concerns about the potential misuse of these images, including the creation of misleading content such as fake news and propaganda. This study investigates the effectiveness of using advanced \acp{vlm} for synthetic image identification. Specifically, the focus is on tuning state-of-the-art image captioning models for synthetic image detection. By harnessing the robust understanding capabilities of large \acp{vlm}, the aim is to distinguish authentic images from synthetic images produced by diffusion-based models. This study contributes to the advancement of synthetic image detection by exploiting the capabilities of visual language models such as \acs{blip}-2 and ViTGPT2. By tailoring image captioning models, we address the challenges associated with the potential misuse of synthetic images in real-world applications. Results described in this paper highlight the promising role of \acp{vlm} in the field of synthetic image detection, outperforming conventional image-based detection techniques. Code and models can be found at \href{https://github.com/Mamadou-Keita/VLM-DETECT}{https://github.com/Mamadou-Keita/VLM-DETECT}
\end{abstract}
\acresetall
\section{Introduction} \vspace{-3mm}
\label{sec:intro}
Fast advances in text-to-image generation models have transformed the landscape of synthetic image creation, enabling users to breathe life into their textual descriptions by generating high-quality visual representations. However, these technological breakthroughs have raised significant concerns regarding the potential misuse of synthetic images for malicious purposes, including the spread of disinformation and propaganda. To safeguard the integrity of information and shield society from the harmful effects of synthetic media, it has become imperative to develop robust techniques to detect and differentiate between real and synthetic images.

In the realm of image generation, various generative models, such as \acp{gan} and diffusion models, have paved the way for the creation of visually authentic images that are almost indistinguishable from real images. Early breakthroughs in synthetic image creation were achieved through the introduction of \acp{gan}~\cite{goodfellow2014generative}. Despite their utility, \acp{gan}-based approaches have been supplemented by diffusion models like DALL-E~\cite{ramesh2021zero}, GLIDE~\cite{nichol2021glide}, Midjourney\footnote{\href{https://www.midjourney.com/}{https://www.midjourney.com/}}, Imagen~\cite{saharia2022photorealistic}, and \ac{sd}~\cite{rombach2022high}, which progressively remove noise from a signal to produce high-quality images~\cite{rombach2022high,saharia2022photorealistic}.

One of the critical challenges associated with these advancements is the development of effective detection methods for synthetic images. Traditional detection techniques, effective against older generative models, face limitations when dealing with the latest diffusion-based architectures and advanced \ac{gan} models. A multitude of research efforts have focused on addressing this concern. For instance, researchers such as in~\cite{corvi2023detection, ricker2022towards} have pointed out the shortcomings of existing detectors in handling these novel generative models, emphasizing the necessity for innovative detection techniques. Authors in~\cite{sha2022fake, coccomini2023detecting} proposed the use of ResNet and other architectures as binary classifiers to distinguish between real and generated images. They further extend these techniques to multimodal detectors by leveraging both image content and accompanying text prompts. In addition, unique properties of diffusion-generated images were explored in~\cite{wang2023dire}, introducing \ac{dire}. Lorenz {\it et al.}~\cite{lorenz2023detecting}  presented multiLID, a technique using local intrinsic dimensionality to estimate densities in \ac{cnn} feature spaces, effectively detecting diffusion-generated images. A significant contribution is \ac{sedid} in~\cite{ma2023exposing}, that leverages diffusion patterns' unique properties and concept of $(t, \delta\text{-error})$ during each diffusion step, to detect real images from those generated by diffusion-based models, enhancing detection accuracy across generative models.

With the emergence of large-scale \acp{vlm}, the field of artificial intelligence has witnessed a transformative evolution. These models, at the intersection of  \ac{nlp} and \ac{cv}, bridge the gap between textual and visual data, thus revolutionizing machine understanding. Prominent examples include CLIP, Flamingo, \ac{blip}-2, Visual-BERT, and ViTGPT2. Within the research community, there is a significant body of work on vision-language tasks, which require the joint processing of information from visual and linguistic modalities to answer complex questions. For example, \ac{vqa} takes an image and a question about the image as input, and outputs the answer to the question. Image captioning, on the other hand, takes an image as input and outputs a natural language description of the image.

\begin{figure*}[t!]
\centering  
\includegraphics[width=.93\linewidth]{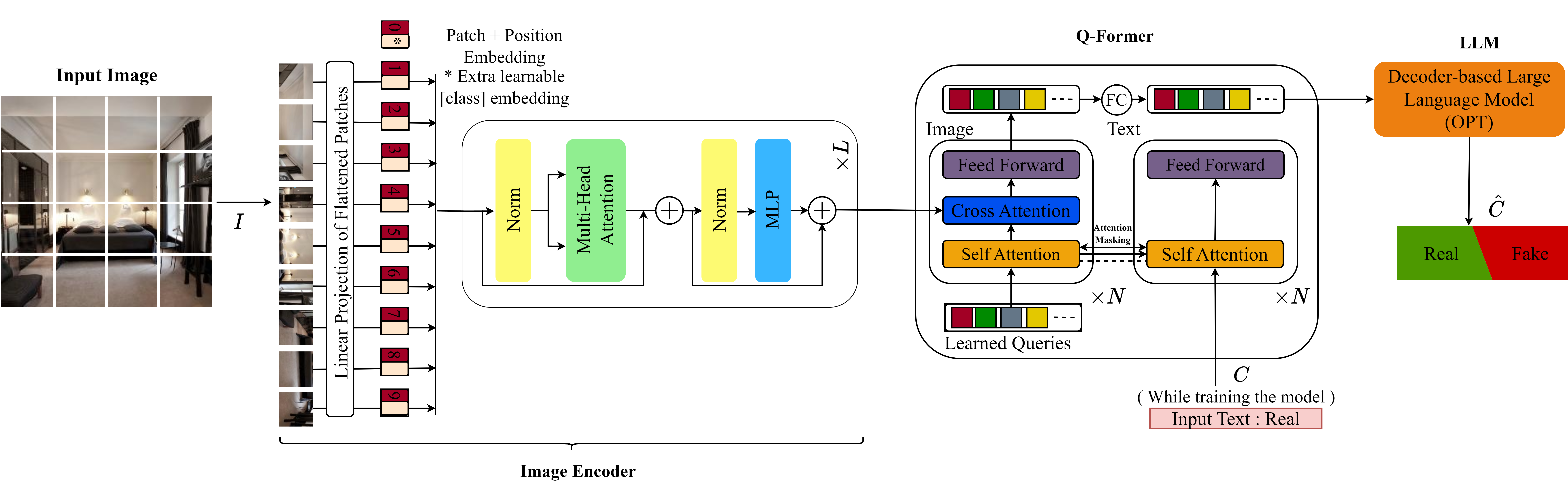}
    \caption{Proposed methodology for synthetic image detection based on the \acs{blip}-2 architecture.} \vspace{-4mm}
  \label{fig:architecture}
\end{figure*}

In this paper, we introduce an innovative approach for synthetic image detection, leveraging state-of-the-art \acp{vlm}. Traditional approaches rely on complex feature extraction processes with \acp{cnn} or \ac{vit} networks for image classification. In contrast, our approach reframes the problem as image captioning, labeling authentic images as "true" and synthetic ones as "fake." This capitalizes on \acp{vlm}' textual content generation skills to distinguish between the two. Our primary focus is on detecting diffusion-generated images, a challenge distinct from previous generative methods like \acp{gan}. Standard binary classifiers struggle to generalize to new diffusion-generated images. To the best of our knowledge, this study pioneers the concept of treating binary classification as image captioning, harnessing cutting-edge \acp{vlm}. Instead of binary classification, we emphasize the potential of using \acp{vlm} like \ac{blip}-2 and ViTGPT2 to create informative captions indicating class membership. In summary, our work makes a threefold pivotal contribution:
\begin{itemize}
\item Reconceptualizing binary classification as image captioning: We redefine binary classification as an image captioning task, harnessing the capabilities of \acp{vlm}.
\item Revealing the potential of \acp{vlm} in synthetic image detection: We shed light on the vast potential of \acp{vlm} in the realm of synthetic image detection, showcasing their robust generalization capabilities, even when faced with previously unseen diffusion-generated images.
\item Empirical validation of enhanced detection: Through rigorous and comprehensive experiments, we substantiate the effectiveness of our proposed approach, particularly in the context of detecting diffusion-generated images.
\end{itemize}
The rest of this paper is structured as follows.   Section~\ref{sec:proposedmethod} describes the proposed \ac{vlm}-based approach for synthetic image detection task. Then, the performance of the proposed approach is assessed and analysed in Section~\ref{sec:results}. Finally, Section~\ref{sec:conclusion} concludes the paper.

\section{Synthetic Image Detection with VLMs}
\label{sec:proposedmethod}

In this section, we present our approach for addressing synthetic image detection using large \acp{vlm}. Our approach re-frames the binary classification task into an image captioning task, leveraging cutting-edge models, including \ac{blip}-2 and ViTGPT2. Fig.~\ref{fig:architecture} provides a visual representation of our methodology built around the \ac{blip}-2 architecture.

\begin{table*}[!ht]
\centering
\caption{Results of different methods trained on \acs{ldm} and evaluated on different testing subsets. We report ACC (\%) / F1-Score (\%).
}
\label{tab:ResultsTable}
{\begin{adjustbox}{max width=0.95\textwidth}
\begin{tabular}{@{}lccccccccc@{}}
\toprule
\multirow{2}{*}{Method} & \multicolumn{8}{c}{Testing Subset} & \multirow{2}{*}{\begin{tabular}[c]{@{}c@{}}Avg\\ (\%)\end{tabular}} \\ \cmidrule(lr){2-9}
       & LDM$^\star$    & ADM$^\oplus$    & DDPM$^\oplus$   & IDDPM$^\oplus$  & PNDM$^\oplus$   & SD v1.4$^\star$ & GLIDE$^\star$  \\ \cmidrule(r){1-1} \cmidrule(l){10-10} 
ResNet50 & 99.92 / 99.92      &  72.33 / 61.83     &  75.26 / 67.21    &   88.96 / 87.61    &  77.20 / 70.52     &   75.47 / 67.57    &    73.10 / 63.28   &        &    80.32 / 73.99   \\
Xception & \bf 99.96 / 99.96       &  52.05 / 7.98    &  58.60 / 29.41    &   54.62 / 16.99     &  60.01 / 33.43    &   63.84 / 43.41   &     58.92 / 30.35 &    & 64.00 / 37.36  \\ 
\acs{deit} & 99.83 / 99.83       &  50.40 / 2.01    &   50.18 / 1.17   &  50.14 / 1.01   &    56.25 / 22.54   &   96.02 / 95.86  &    98.15 / 98.11  & &  71.56 / 45.79   \\
ViTGPT2  & 99.40 / 99.40 & 70.84 / 59.21 & 69.60 / 56.72  & 84.08 / 81.20 & 95.40 / 95.22 &  \bf 99.54 / 99.55  &  \bf 99.27 / 99.27 &  & 88.30 / 84.37\\ 
\acs{blip}-2  & 99.12 / 99.13 & \bf 85.24 / 82.97 & \bf 98.47 / 98.47 & \bf 97.02 / 96.97 & \bf 99.22 / 99.23 &      77.68 / 71.79  &    97.09 / 97.05  &        & \bf 93.41 / 92.23 \\ \bottomrule
\end{tabular} \vspace{-8mm}
\end{adjustbox}}
 \begin{flushleft}
\scriptsize $^\star$ Text-To-Image diffusion-based model.  $^\oplus$ Unconditional diffusion-based model. \vspace{-8mm}
\end{flushleft}
\end{table*}

\subsection{Synthetic Image Detection}
\label{sec:SyntheticImageDetection}

Synthetic image detection is generally a binary image classification task. Its main objective is to develop a model \(\mathcal{M}\) that learns a function \(f : \mathbf{I} \rightarrow \mathbf{Y}\) from the training set \(D = \{({I}_i, {y}_i) | 1 \leq i \leq n\}\), where \({I}_i \in \mathbf{I} = \mathbb{R}^{d \times d}\) is an image, and \(y_i \in \mathbf{Y} = \{0, 1\}\) represents the class labels. Here, \(0\) indicates images captured from the real world, and \(1\) indicates generated images. The model's objective is to predict the class \(\hat{y}\) for an input image \({I}\) as follows: 
\begin{equation}
    \hat{y} = f_{\theta}({I}),
\end{equation}
where \(\hat{y}\) is the predicted class label (0 or 1), and \(\theta\) represents the model parameters. \vspace{-3mm}
\subsection{Fine-Tuning}
\label{sec:FineTuning}

In the previous subsection, we introduced synthetic image detection from a general perspective and will now proceed to re-frame it as an image captioning problem, harnessing the capabilities of \acp{vlm}. Instead of treating synthetic image detection as a traditional binary classification task, we will employ \acp{vlm} to generate descriptive captions for each image. The \acp{vlm} will learn to produce captions that capture the essence of the image, and these captions will serve as indicators of the image's authenticity. Specifically, the model will generate captions that fall into one of two categories: "real" or "fake."

Let's define this fine-tuning process mathematically. We have a \ac{vlm} \(\mathcal{M}\) with parameters \(\theta\), which takes an image \(I\) as input and generates a caption \(\hat{C}\). We denote this process as:

\begin{equation}
\hat{C} = \mathcal{M}_{\theta}(I).
\end{equation}

The caption \(\hat{C}\) is a human-like label (textual description) of the image, and it will be used to distinguish the "real" image from the "fake".

To tune the model for this task, we have a dataset \(D = \{(I_i, C_i) | 1 \leq i \leq n\}\), where \(I_i\) is the \(i\)-th image, and \(C_i\) is the ground truth caption whether the image is "real" or "fake". The model's objective is to minimize a suitable loss function \(\mathcal{L}\) over this dataset:

\begin{equation}
\theta^* = \arg\min_\theta \sum_{i=1}^{n} \mathcal{L}(\mathcal{M}_{\theta}(I_i), C_i).
\end{equation}

Once fine-tuned, the \ac{vlm} will be capable of generating captions that help categorizing images into the desired class, providing a more detailed and expressive approach to synthetic image detection.
In the following subsections, we will delve into the details of the fine-tuning process of two \acp{vlm}, ViTGPT2 and \Ac{blip}-2.

\subsection{ViTGPT2}
\label{sec:vitgpt2}
ViTGPT2\footnote{\label{myfootnote}\href{https://ankur3107.github.io/blogs/the-illustrated-image-captioning-using-transformers/}{https://ankur3107.github.io/blogs/the-illustrated-image-captioning-using-transformers/}} is a vision encoder-decoder model that belongs to the family of models with a pre-training objective that involves directly fusing visual information into the layers of a language model decoder. As the name suggests, ViTGPT2 adopts the \ac{vit} as the image encoder and the \ac{gpt}-2 model as the language model. For our novel synthetic image detection approach, we instantiated the model using the image-to-text model class \texttt{VisionEncoderDecoderModel} where cross-attention layers are automatically added to the decoder. During fine-tuning with Hugging Face's \texttt{Seq2SeqTrainer}, only these added layers are tuned. ViTGPT2 leverages the strengths of a transformer-based vision model (encoder) and a pretrained language model (decoder) to process visual information effectively.

\vspace{-3mm}
\subsection{BLIP-2}
\label{sec:blip2}

\begin{figure}[t!]
\centering
\includegraphics[width=\linewidth]{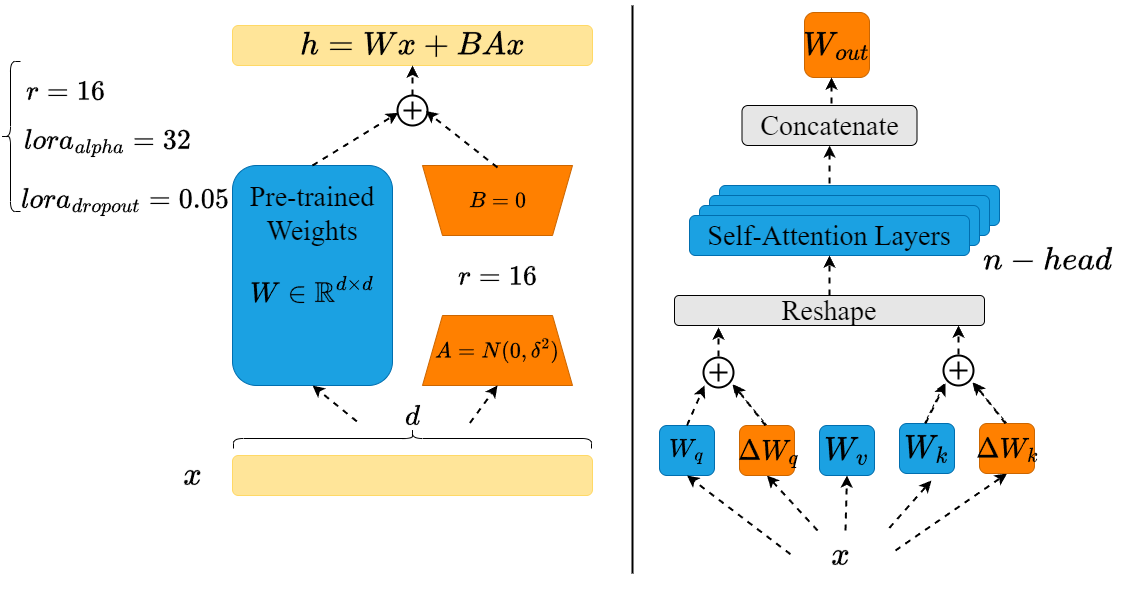}
    \caption{\acs{blip}-2 fine-tuning with \ac{lora}.}
  \label{fig:lora} \vspace{-5mm}
\end{figure}

\Ac{blip}-2~\cite{li2023blip} is a method that improves vision-language alignment using a \ac{q-former} as a bridge. \Ac{q-former} extracts relevant visual information from frozen image encoders and provides it to a frozen language model. The method undergoes two pre-training phases: one for learning visual representations for text and another for training the \ac{q-former} to provide interpretable visual input to the language model. Thus, the smaller matrices are trained to learn task-specific information using supervised learning.
Fig.~\ref{fig:lora} provides a visual overview of fine-tuning the \ac{blip}-2 model using the \ac{lora} technique. During fine-tuning, we keep the pretrained \ac{blip}-2 model weights (\(W\)) frozen and inject a pair of trainable rank decomposition matrices (\(A\) and \(B\)) into each self attention layer, enabling us to optimize and adapt these layers to our new detection task without incurring excessive computational overhead. Thus, the smaller matrices are trained to learn task-specific information using supervised learning. This method is commonly referred to as \ac{lora}~\cite{hu2021lora}, which falls under the re-parameterization category of \ac{peft} methods. Specifically, the update of BLIP-2 pre-trained weight matrix $W$ is constrained by a low-rank decomposition $W + \Delta W = W + BA$. Note that both $W$ and $\Delta W = BA$ are multiplied by the same input $x$, and their respective output vectors are summed coordinate-wise. For $h = Wx$, the modified forward pass yields:
\begin{equation}
    h = Wx + \Delta W x = Wx + BAx.
\end{equation}

It is worth highlighting that, among various attention weights, we adapt only two types: \(W_q\), and \(W_k\).

\vspace{-3mm}
\section{Results}
\label{sec:results}


\begin{figure*}
    \centering
    \begin{subfigure}[b]{0.21\textwidth}
        \includegraphics[width=.8\linewidth]{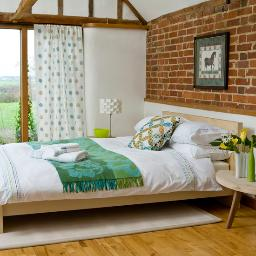}
        \caption{Real - $0|0|0|0|0$ \hspace{8mm} }
    \end{subfigure}
    \begin{subfigure}[b]{0.21\textwidth}
        \includegraphics[width=.8\linewidth]{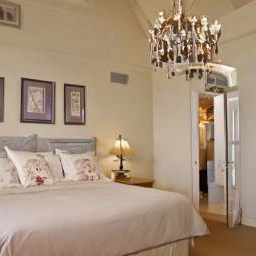}
        \caption{Fake (ADM) - $0|0|0|0|1$ \hspace{15mm}}
    \end{subfigure}
    \begin{subfigure}[b]{0.21\textwidth}
        \includegraphics[width=.8\linewidth]{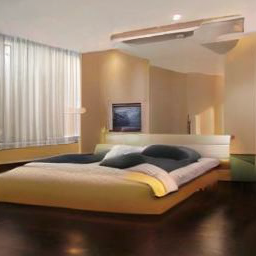}
        \caption{Fake (LDM) - $1|1|1|1|1$ \hspace{5mm}}
    \end{subfigure}
    \begin{subfigure}[b]{0.21\textwidth}
        \includegraphics[width=.8\linewidth]{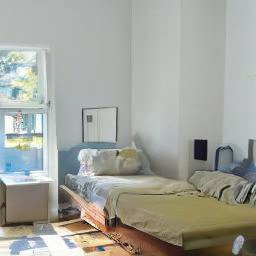}
        \caption{Fake (DDPM) - $0|0|0|1|1$ \hspace{5mm}}
    \end{subfigure}
    \begin{subfigure}[b]{0.21\textwidth}
\includegraphics[width=.8\linewidth]{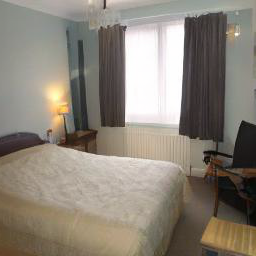}
        \caption{Fake (IDDPM) - $1|0|0|0|1$}
    \end{subfigure}
    \begin{subfigure}[b]{0.21\textwidth}
        \includegraphics[width=0.8\linewidth]{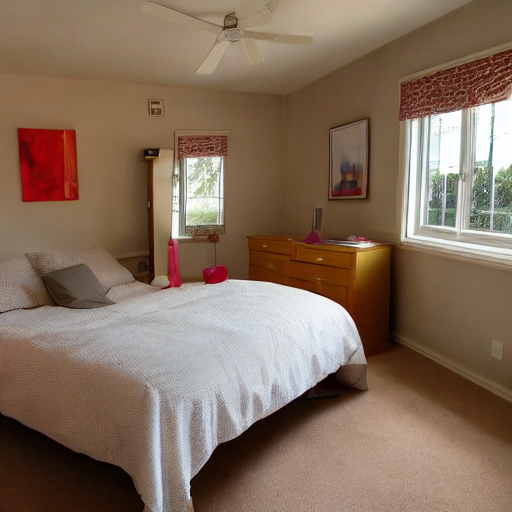}
        \caption{Fake (SD) - $0|1|0|1|1$}
    \end{subfigure}
    \begin{subfigure}[b]{0.21\textwidth}     \includegraphics[width=.8\linewidth]{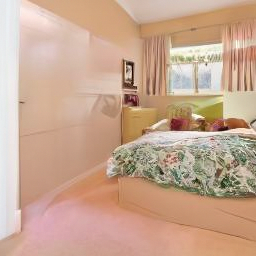}
        \caption{Fake (PNDM) - $0|0|0|1|1$}
    \end{subfigure}
     \begin{subfigure}[b]{0.21\textwidth}     \includegraphics[width=.8\linewidth]{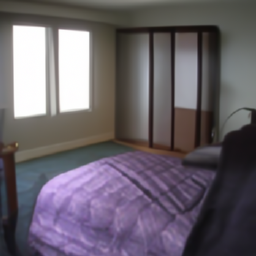}
         \caption{Fake (GLIDE) - $0|0|1|1|1$}
     \end{subfigure}  
    \caption{Each sub-figure is a random image from a testing set, labeled below. The 5-digit binary code shows results from ResNet, Xception, \acs{deit}, ViTGPT2, and \acs{blip}2 models, where '0' means real and '1' means fake. All generated images are considered fake.}
    \label{fig:Visual} \vspace{-4mm}
\end{figure*}

\subsection{Experimental Setup}
\textbf{Dataset.} To effectively assess the performance of large-scale \acp{vlm} in synthetic image detection, we utilized an existing dataset introduced in the work by Ricker {\it et al.}~\cite{ricker2022towards}. In this dataset, all the real images are sourced from the \ac{lsun} Bedroom dataset~\cite{yu2015lsun}. We gathered a collection of images generated by five distinct diffusion models that were trained on the \ac{lsun}-Bedroom dataset. Among these models, four subsets of generated images (namely, ADM~\cite{dhariwal2021diffusion}, DDPM~\cite{ho2020denoising}, iDDPM~\cite{nichol2021improved}, and PNDM~\cite{liu2022pseudo}) were produced by unconditional diffusion models. The fifth subset, referred to as LDM~\cite{rombach2022high}, was generated using text-to-image diffusion models. To assess the adaptability of these models to generative text-to-image models, we expanded the set of text-to-image generation models by incorporating two additional models: \ac{sd}~\cite{rombach2022high} and GLIDE~\cite{nichol2021glide}. For each of these models, we generated a test set consisting of 10,000 images. The prompt used to generate these images was "A photo of a bedroom". In our experiment, we use a training set of 40,000 images and a testing set of 10,000 images for the LDM subset. For other generative models, we only use a testing set of 10,000 images. It's worth noting that the real images used for testing were consistent across all the testing subsets. \\\\
{\textbf{Evaluation metrics.}} Following the convention of previous detection methods~\cite{ma2023exposing,coccomini2023detecting,lorenz2023detecting}, we also report the accuracy (acc) and F1 score (f1-score) in our experiments.  Accuracy (acc) measures the proportion of correct predictions, with higher scores indicating better performance. While F1-score balances precision and recall, offering a single measure of a model's accuracy in identifying positive instances while minimizing false results. \\\\ 
{\textbf{Baselines.}} To comprehensively evaluate and compare our proposed approach, we tailored several prominent models, namely ResNet~\cite{he2015deep}, Xception~\cite{Chollet_2017_CVPR}, and \ac{deit}~\cite{touvron2021training}, to establish baseline benchmarks. We opted for these models due to their extensive usage and outstanding performance in related tasks. To formulate our baseline models, we fine-tuned these architectures by substituting their final \ac{fc} layers with a novel  \ac{fc} layer featuring a single neuron dedicated to discerning the authenticity of images. These models were initialized with pre-trained weights gleaned from the ImageNet dataset~\cite{deng2009imagenet}, thereby harnessing the knowledge encoded in their learned representations. Employing these custom-tailored baselines enables us to assess the efficacy of our proposed approach when compared with well-established and widely-adopted image classification architectures. \\\\
{\textbf{Implementation details.}} In our experiments, we leveraged the PyTorch deep learning framework on a Windows computer equipped with a 16GB NVIDIA RTX A4500 GPU. Our study used baseline models obtained from their publicly available repositories, which we fine-tuned to align with our specific experimental setup. For \acs{blip}-2 model tuning, we opted for the Adam~\cite{kingma2014adam} optimizer with default settings, a learning rate of 5e-5, and training duration of 20 epochs. During training, we configured the \ac{peft} module, in particular \ac{lora}, with the followings params: a \textit{rank}  of 16, \textit{lora\_alpha} set to 32, \textit{lora\_dropout} at 0.05, and a \textit{batch\_size} of 32. Regarding ViTGPT2 model, we followed a procedure outlined by the author, as detailed in provided note\footref{myfootnote}. To ensure a fair comparison, we retrain all baselines on our training dataset, instead of using their published versions directly. \vspace{-2mm}
\subsection{Results and Analysis}
In this section, we present the comprehensive results of our experimental evaluation of the effectiveness of large \acp{vlm} in detecting synthetic images from various generative models. Our study aimed to explore the performance of different \acp{vlm} across various test subsets and provide insights into their ability to discern synthetic visual content.
Table~\ref{tab:ResultsTable} presents the accuracy and F1-Score performance of the examined models, which were trained on the LMD dataset and subsequently tested on seven distinct image generation models.

In all test subsets, models demonstrated variable levels of accuracy in detecting synthetic images. ResNet50 demonstrated consistently high accuracy, ranging from 72.33\% to 88.96\%, in all subsets tested. In contrast, Xception performed relatively poorly, with accuracy ranging from 52.05\% to 63.84\%. \ac{deit} showed an interesting trend, achieving an accuracy of 96.02\% on the SD v1.4 subset, whereas its accuracy was significantly lower on other subsets, indicating potential challenges in certain synthetic image detection scenarios. Specifically, \ac{deit} achieved good performance in text-to-image generative models (LDM, SD v1.4, GLIDE), indicating strong generalization to unseen models from the same subcategory as the one used for training. ViTGPT2 performed competitively, with accuracy rates ranging from 69.60\% to an impressive 99.54\%. \ac{blip}-2 stood out in particular, posting outstanding accuracy rates on all subsets, ranging from 77.68\% to 99.22\%. This indicates its robustness in detecting various types of synthetic content.

Our findings reveal that the synthetic image detection performance of \acp{vlm} is influenced by factors such as model architecture and the complexity of the synthetic image generation technique. Models such as \ac{blip}-2 and VitGPT2, which incorporate both visual and linguistic information, show more consistent performance across subsets as it can be seen in Fig.~\ref{fig:Visual}, suggesting the potential benefits of using a combined vision-language approach.

These findings prompt a discussion of the implications of \acp{vlm} in real-world applications, particularly in security and content moderation systems. The robust performance of models such as \ac{blip}-2 raises questions about their ability to reliably detect sophisticated synthetic content, which could have implications for the identification of misinformation and visual fake media. \vspace{-3mm}

\section{Conclusion}
\label{sec:conclusion}

In this paper, we have presented a new approach to the synthetic image detection task by exploiting the capabilities of state-of-the-art \acp{vlm}. By re-framing the challenge of binary classification as an image captioning task, we showed the potential of models such as ViTGPT2 and \ac{blip}-2 to transcend traditional boundaries and provide new perspectives in image authenticity assessment. Our methodology relies on the ability of these models to fuse visual and textual information, enabling them to generate captions that encapsulate the essence of both real and synthetic images. The results of our experiments have shown that this approach outperforms conventional classification techniques. Our results have implications beyond the immediate task of image classification. They highlight the potential of \acp{vlm} to solve complex problems that benefit from multimodal understanding. As these models evolve, the boundaries of their applications expand, promising to revolutionize various fields that rely on nuanced perception.

\bibliographystyle{IEEEbib}
\bibliography{strings,refs}

\end{document}